# Quantifying Relational Exploration in Cultural Heritage Knowledge Graphs with LLMs: A Neuro-Symbolic Approach

Mohammed Maree

Department of Information Technology, Faculty of Information Technology, Arab American University, Jenin P.O. Box 240, Palestine

**Abstract**
This paper introduces a neuro-symbolic approach for relational exploration in cultural heritage knowledge graphs, leveraging Large Language Models (LLMs) for explanation generation and a novel mathematical framework to quantify the *interestingness* of relationships. We demonstrate the importance of *interestingness* measure using a quantitative analysis, by highlighting its impact on the overall performance of our proposed system, particularly in terms of precision, recall, and F1-score. Using the Wikidata Cultural Heritage Linked Open Data (WCH-LOD) dataset, our approach yields a precision of 0.70, recall of 0.68, and an F1-score of 0.69, representing an improvement compared to graph-based (precision: 0.28, recall: 0.25, F1-score: 0.26) and knowledge-based baselines (precision: 0.45, recall: 0.42, F1-score: 0.43). Furthermore, our LLM-powered explanations exhibit better quality, reflected in BLEU (0.52), ROUGE-L (0.58), and METEOR (0.63) scores, all higher than the baseline approaches. We show a strong correlation (0.65) between interestingness measure and the quality of generated explanations, validating its effectiveness. The findings highlight the importance of LLMs and a mathematical formalization for *interestingness* in enhancing the effectiveness of relational exploration in cultural heritage knowledge graphs, with results that are measurable and testable. We further show that the system enables more effective exploration compared to purely knowledge-based and graph-based methods.

## 1. Introduction

The digitization of cultural heritage artifacts and historical records has generated a vast amount of knowledge encoded in the form of interconnected knowledge graphs (KGs) [1, 2]. Unlocking meaningful insights from these KGs requires more than simple keyword searches [3]. Relational search – the discovery of connections between entities – is crucial for this goal, enabling to uncover complex patterns and narratives that might remain hidden in raw data. Traditional knowledge-based approaches rely on predefined patterns, limited explanation generation and still requires manual engineering [4, 5]. These methods can lack the flexibility and adaptability necessary to uncover the full range of nuanced relationships inherent in complex cultural heritage data [6]. They also fail to truly reflect what might be deemed as *interesting* to an end-user, particularly users with varying degrees of expertise in the domain [7]. This paper introduces a novel neuro-symbolic approach to relational search in cultural heritage KGs. Our proposed method leverages the capabilities of Large Language Models (LLMs) to enhance both the discovery and explanation processes. The motivation for this research comes from the observation that relational search requires not only the discovery of relationships, but also a high-level human interpretable explanation that is contextually relevant. We aim to go beyond simple graph traversal algorithms, such as Breadth-First Search (BFS) and Depth-First Search (DFS), or pre-defined inflexible SPARQL queries, and use LLMs to help in both the generation and explanation of the underlying data. Specifically, we aim to:

- Generate diverse and contextually relevant explanations through the exploitation of LLMs; generating explanations that go beyond predefined templates, and adapting to different semantic contexts to provide a deeper understanding of the relationships uncovered.
- Personalize explanations through taking the user context into account to meet individual preferences and knowledge levels.

- Quantify *interestingness* with a mathematical framework, where our proposed system can guide users towards potentially surprising and valuable relationships. This process is more informed compared to pure exploration of the data through faceted search.

Further, we introduce a new formal framework, where we treat *interestingness* as a measurable concept, mathematically defined and computed, thereby making our approach data-driven. The framework is a crucial ingredient in our approach, since it enables the system to be guided towards the discovery of potentially interesting relationships, and for the LLMs to generate the explanations. We validate this methodology through comprehensive experiments on a rich cultural heritage KG, demonstrating its effectiveness in uncovering non-trivial relationships and providing insightful explanations.

The rest of this article is organized as follows: In Section 2, we present related work on existing relational search methods, highlighting their limitations. Section 3 introduces our proposed methodology, detailing the connection discovery process, our mathematical framework for *interestingness* scoring, and how we leverage LLMs for explanation generation. Section 4 provides an overview of the experimental setup, including dataset details, implementation specifics, and evaluation procedures. We present and discuss the experimental results in Section 5, showing quantitative measures of performance. Finally, Section 6 summarizes our contributions, and outlines potential avenues for future research.

## 2. Related Work

The challenge of uncovering meaningful connections within knowledge graphs (KGs) has spurred a diverse range of research efforts [2, 5, 8-24]. In this section, we explore and discuss existing techniques, grouping them into relevant categories with an emphasis on their strengths and limitations. We organize these techniques into knowledge-agnostic graph traversal methods, knowledge-based approaches, explainable AI in knowledge graphs, and the emerging use of LLMs in knowledge graph reasoning.

### 2.1 Knowledge-Agnostic Graph Traversal

These methods operate on the structural properties of the KG, treating it as a network of interconnected nodes and edges. They aim to find relationships between entities without relying on the semantic meaning of the data itself. This makes them generalizable but also prone to generating a lot of irrelevant results [8]. In addition, these approaches are based on simple graph traversal methods. For example, using fundamental algorithms such as the Breadth-First Search (BFS) and Depth-First Search (DFS), relationships between nodes are systematically explored in KGs [9]. BFS starts from a source node and visits all its neighbors before moving to the next level, while DFS explores as far as possible along each branch. Although these approaches can find all reachable nodes or paths between nodes, they are computationally expensive in large KGs and frequently generate numerous paths that are semantically trivial. They often produce a combinatorial explosion of results that are uninteresting or obvious to a user, requiring manual post-processing and filtering [10]. Other algorithms like Dijkstra's and A*, find the shortest path between two entities based on edge weights or hop counts. Methods like WiSP [11], can be used to find the most direct paths but still do not capture semantic relatedness or relevance to the user's context. Although these algorithms can find the most direct connection between entities, that path may not always be the most relevant, surprising, or informative for relational exploration, and might be too simplistic to capture the nuances of the relationships [12]. On the other hand, Random Walk and Path Ranking techniques simulate a random walk across the graph to compute node importance and explore connectivity. Methods like Personalized PageRank use random walks to identify relevant results for specific nodes, however, while these methods are scalable and more efficient than other graph traversal methods, they do not typically capture the rich semantics of relationships and user context [13]. They also can produce many irrelevant and uninterpretable paths. Accordingly, we would like to point out that, while efficient for exploring the structure of KGs, these knowledge-agnostic methods fall short in addressing the core problem of relational search. They are not guided by any form of domain understanding, and therefore often generate a plethora of irrelevant results and provide limited explanations that lack context. The paths are not always human understandable, and do not provide deep explanations of the relationship discovered.

### 2.2 Knowledge-Based Approaches

These techniques leverage domain-specific knowledge, often encoded in ontologies, rules, and schemas, to guide the relational exploration, but at the expense of flexibility and manual engineering [2, 14]. These techniques aim to mitigate some of the limitations of purely graph-based approaches by incorporating more structure and semantics.

For instance, ontology-based search methods use formal ontologies (e.g., CIDOC CRM) to reason about relationships between entities [2]. While ontologies provide a structured vocabulary for representing knowledge, they are mostly designed for structured data retrieval and do not provide the flexibility or ability to generate natural language explanations that are tailored to a user. They are often more suitable for querying structured data instead of providing explanations. On the other hand, SPARQL-based approaches use the SPARQL query language to specify precise patterns of interest [15]. SPARQL offers a powerful way to extract complex data from RDF graphs. However, such approaches often rely on manually created queries and pre-defined templates for explanations, which are brittle and hard to adapt to varying user needs, or different contexts. In a similar line of research, rule-based methods use predefined rules to infer and discover new relations [5]. Although they can provide some level of domain-specific reasoning, they are generally limited by the scope of the rules and their lack of adaptability. Moreover, these rules need to be specified by domain experts which increases the workload in creating, maintaining and adapting them for different scenarios. These approaches address precision to some extent but are limited by the scope of predefined rules and templates, as well as their lack of personalization. As noted in our introduction, they fail to capture the nuances of relationships and are not adaptable to varying user contexts [16, 17]. Accordingly, these approaches lack the ability to provide flexible explanations for a varied number of users in real-world application contexts and settings.

## 2.3 Explainable AI in Knowledge Graphs

Explainable AI (XAI) in the context of KGs aims to provide human-understandable explanations for the relationships discovered in the graphs. However, this area is still in early stages and faces some limitations. For instance, in Path-Based explanation methods, like RelFinder [18], connections are visualized as graph paths or subgraphs connecting two or more entities. However, these paths and subgraphs might be hard to interpret and understand, especially for users who are not domain experts, or have no technical background. These methods also do not provide natural language explanations [19]. Other Subgraph-Based explanation methods extract a portion of the knowledge graph to create an explanation that provides more context than simple paths [20]. However, subgraphs still require some effort for interpretation by end users, as it is not in natural language. In addition, as reported in [21], such methods require substantial computational resources, leading to extensive utilization of approximation techniques and other optimization approaches. Other Graph Neural Networks (GNNs) have been also applied for reasoning and link prediction [22, 23]. While GNNs can learn complex patterns and produce good results on link prediction, they are also very complex to interpret, as they often operate as a black boxes and are not suitable for generating natural language explanations to the end-user [24]. As reported in a recent review in [25], there are four open problems, including robustness, interpretability, pretraining and complex structure modeling that still form a major challenge for GNNs and hinder their exploitation in real-world and practical application domains, such as the CH domain. Therefore, while these methods try to address the lack of transparency of AI systems, they still struggle to create rich, human-readable, and context-aware explanations that go beyond just graph paths or subgraphs. They often lack the ability to tailor explanations to the context or knowledge level of the user and require a level of technical expertise.

## 2.4 Large Language Models in Knowledge Graph Reasoning

The use of Large Language Models (LLMs) in KG applications is an emerging field. Although LLMs are showing great potential in many other areas, their integration with relational search in KGs is still in early stages. For instance, LLMs can answer questions based on the information available in KGs. They can extract relevant entities, use reasoning and language understanding, and provide answers to natural language questions. Moreover, they can be trained on KG embeddings for performing link prediction, by learning representations of nodes and relationships. However, these approaches often focus on the accuracy and efficiency of link prediction but don't focus on generating high-quality explanations. On the other hand, recent research attempts have explored methods for fusing neural networks with symbolic knowledge representations for explainable AI. These approaches including the work of Díaz-Rodríguez et al.[26] and Arrieta et al. [27] combine a neural network for image classification with a knowledge graph to guide the detection of object-parts. These approaches use a knowledge graph and a neural network in a fused way to train the neural network, where the knowledge graph is used as a form of regularization. However, they do not make use of an LLM for generating natural language explanations, nor do they use a user context for tailoring explanations. They also do not focus on relational exploration, but rather on improving the performance and explainability of object classification through the use of a knowledge graph. As such, and while

LLMs have demonstrated new potentials in these areas, their integration into relational search is not fully explored, particularly in generating human-understandable explanations of connections and not just retrieving data, and guiding the search towards more interesting connections. Also, LLMs are typically not used with a mathematical framework for guiding the explanations, as proposed in this work, or use a user context for personalizing results.

## 3. Proposed Methodology

Our approach is a neuro-symbolic framework that combines elements of knowledge-based systems with the power of LLMs, augmented by a novel mathematical framework to guide both the selection of connections and the generation of their explanations. This methodology is designed to enhance the process of relational exploration, enabling users to discover and understand complex, non-obvious connections within a knowledge graph in a personalized way. Unlike approaches focused on tasks such as object classification, our goal is to provide a comprehensive system to generate explanations based on the relationships between entities, rather than focusing only on the properties of each entity. As depicted in Figure 1, this process can be broken down into three main stages, each building upon the previous one: Connection Discovery, *Interestingness* Scoring, and Explanation Generation.

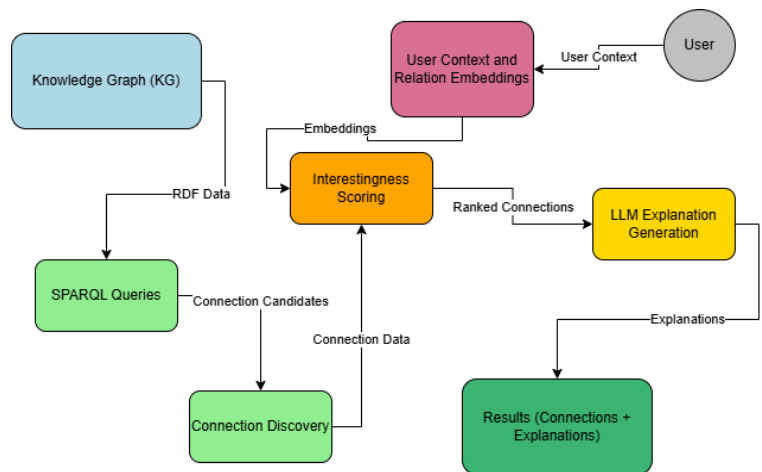

Figure 1. System Architecture Overview

### 3.1 Connection Discovery

This initial stage is dedicated to identifying potential relationship instances between entities within the knowledge graph. This step is not focused on extracting all possible relations between all entities, but rather to extract instances of relations that match pre-defined patterns of connections between entities, that are commonly found in the domain. This helps us to prune down the search space and avoid the problem of a combinatorial explosion of results. To achieve this, we utilize a series of SPARQL CONSTRUCT queries, which leverage the semantic structure of the KG to identify relationship patterns. Instead of producing explanation instances directly at this step, the SPARQL queries generate candidate connections, which can be analyzed using our mathematical model. In addition, Algorithm 1 below details the pseudo-code for discovering connections between entities.

**Algorithm 1: Connection Discovery**
**Input:**
- ⇨ KG: The knowledge graph.
- ⇨ SPARQL_Queries: A set of SPARQL CONSTRUCT queries defining relationship patterns.

**Output:** Connection_Candidates *CC*: A set of candidate relationship instances.
1. *CC* = < > an empty list.
2. For each SPARQL_Query in SPARQL_Queries:
    - o Execute the SPARQL_Query on the KG.
    - o For each result found by the SPARQL_Query:
        - ▪ Create a connection_instance object including:
            * The identifier of the two connected entities

* The type of relationship between them
    * Relevant metadata such as times or additional information
    * A simple textual label for the relationship
  - Add connection_instance to Connection_Candidates.
3. Return *CC.*

Figure 2 provide a detailed illustration of the connection discovery process.

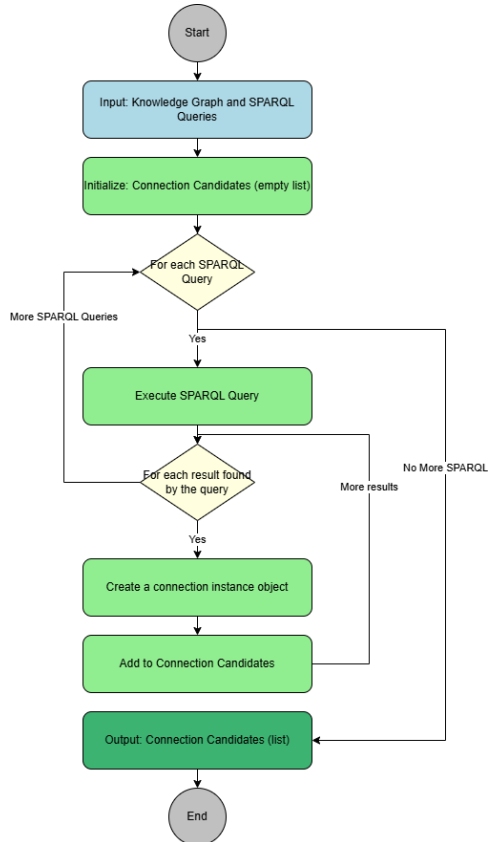

Figure 2. Connection Discovery Process

Specifically, we have identified several common types of relationships between entities, which are encoded in our set of SPARQL queries. For example:
- "person X was born in place Y"
- "person X works in place Y"
- "person X wrote a book about place Y"
- "person X created a painting that depicts place Y"

The results of the queries are used to produce sets of data that contain all the basic information required to compute our interestingness score. Each connection instance will contain:
- entity1_id: The identifier of the first connected entity.
- entity2_id: The identifier of the second connected entity.
- relationship_type: The type of relationship connecting the entities.
- relevant_metadata: Relevant metadata, such as times or additional information (when available).
- explanation_text: A simple textual label describing the relationship. This will be later used by the LLM for generating explanations.

This stage, therefore, provides a foundation for the next stage by generating a set of candidate relationships to explore.

## 3.2 Formalizing *Interestingness* of Relationships

The core of our approach lies in our novel mathematical framework for quantifying the *interestingness* of a given relationship instance. We do so by combining the notions of semantic relatedness within the knowledge graph, and the contextual relevance of the connection to a specific user.

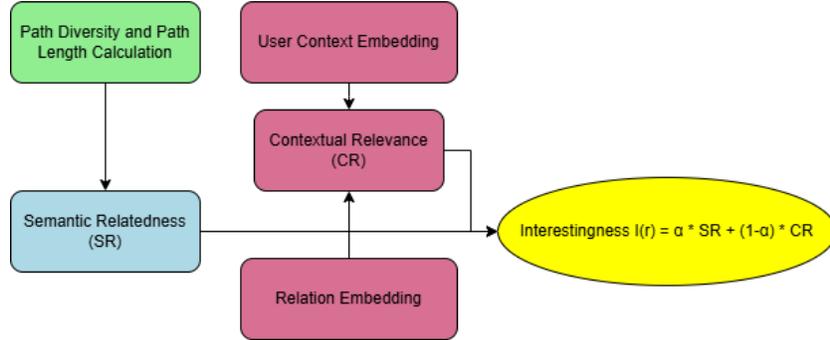

Figure 3. Interestingness Scoring Calculation

As depicted in Figure 3, our measure of interestingness, is given by:

$$I(r) = \alpha * SR(e1, e2) + (1 - \alpha) * CR(r, U)$$

Where:
- $I(r)$: The overall interestingness score of the relationship instance *r* between entities *e1* and *e2*.
- $SR(e1, e2)$: The semantic relatedness score between entities *e1* and *e2*. This quantifies how closely related two entities are in the context of the graph structure.
- $CR(r, U)$: The contextual relevance score of the relationship *r* to the user's context *U*. This takes into consideration the user's preferences, history and expertise.
- $\alpha$: A weight parameter (0 ≤ α ≤ 1) that balances the influence of semantic relatedness and contextual relevance.

The parameter $\alpha$ is used to control the balance between semantic relatedness and the user context. A value of $\alpha$ close to 1 will give more weight to the semantic relatedness measure, and a value closer to 0 will give more weight to the contextual relevance. The parameter $\alpha$ can be customized by the user, or it can be automatically set by the system by using a user profile.

### 3.2.1 Semantic Relatedness (SR)

Semantic relatedness measures the intrinsic relationship between two entities based on the connectivity and diversity of paths linking them in the KG. Our formulation favors shorter, more diverse paths, which often indicate a more meaningful semantic connection. The semantic relatedness score is calculated as follows:

$$SR(e1, e2) = (1 / (|P| + 1)) * \Sigma (1 / dist(pi))$$

Where:
- $P$: The set of all simple paths connecting entity *e1* and *e2* within the KG. A simple path is a path that does not visit a node twice.
- $|P|$: The total number of simple paths connecting *e1* and *e2*.
- $dist(pi)$: The length of the $i-th$ path $pi$, measured as the number of edges traversed.

The term $(1 / (|P| + 1))$ is a normalization factor that takes into account path diversity. The higher the path diversity (the more paths are available), the lower the semantic relatedness score will be.

### 3.2.2 Contextual Relevance (CR)

Contextual Relevance captures how well a given relationship aligns with a user's current needs and preferences. We use a similarity measure between vector embeddings of both the relationship and the user's context. The contextual relevance is computed as the cosine similarity between these embeddings:

$$CR(r, U) = cosine(v(r), v(U))$$

Where:
- $v(r)$: The vector embedding of the textual description of the relationship $r$, generated by a Large Language Model (LLM).
- $v(U)$: The vector embedding of the user's context $U$, also generated by an LLM. The user context $U$ can be a concatenation of different types of information: search history, domain expertise, specified user interests or other available information. The LLM can take this combined context information to generate the embeddings.
- $cosine(v(r), v(U))$: The cosine similarity between the embeddings of $v(r)$ and $v(U)$. The score ranges from -1 to 1, with values closer to 1 indicating a higher degree of contextual relevance.

This step introduces the use of LLMs in the system, by allowing the system to process natural language and create suitable vector embeddings.

### 3.3 Explanation Generation with LLMs

Once the *interestingness* score has been computed, we use an LLM to generate a human-readable natural language explanation. The LLM is prompted with a structured set of information about the connection instances, which includes:
- entity1_description: A concise description of the first entity.
- entity2_description: A concise description of the second entity.
- relationship_type: The type of relationship connecting the two entities.
- interestingness_score: The value of $I(r)$.
- user_context_description: A textual representation of the user's context $U$.

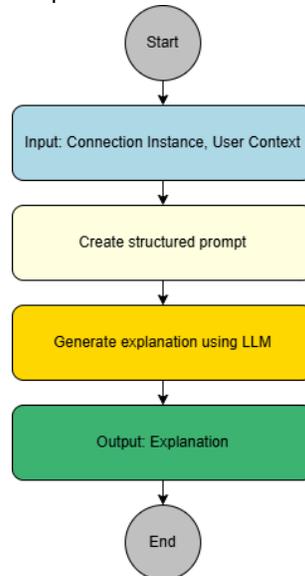

Figure 4. Explanation Generation Process

As shown in Figure 4, the following prompt structure is used to generate explanations, which is given to the LLM:
"*Generate a natural language explanation that connects `entity1_description` and `entity2_description`.
The relationship type is `relationship_type`.
The interestingness score is: `interestingness_score`.
The user context is: `user_context_description`.
Explain this relationship in a way that reflects its interestingness, and the user context. Be specific, and avoid generic statements that could apply to other entities.*"

The LLM uses the provided information to create an explanation that reflects the computed interestingness score and the user-specific context. The LLM is explicitly prompted to create a personalized explanation that highlights the nuances and significance of the relationship. The LLM is explicitly instructed to create non-generic explanations, and to use any other relevant information that may improve the quality of the final explanation. We also provide different types of examples to the LLM in order to facilitate and guide it towards creating high quality explanations.

**Algorithm 2: Explanation Generation**
**Input:**
- Connection_Candidates: A set of candidate relationship instances.
- User Context $U$: A user context vector.
- LLM: The Large Language Model.

**Output:** Connection_Candidates $CC$: The same set of connections, but with the addition of natural language explanation for each instance.

1. For each connection instance r in Connection_Candidates:
   - Compute $SR(e1, e2)$ with the KG, using formula (Eq. 2) (where $e1$ and $e2$ are the entities linked by $r$).
   - Compute $v(r)$ and $v(U)$ vector embeddings using the LLM (Eq. 4).
   - Compute $CR(r, U)$ using $cosine(v(r), v(U))$ (Eq. 5).
   - Compute the $I(r)$ using $\alpha * SR(e1, e2) + (1 - \alpha) * CR(r, U)$ (Eq. 1)
   - Prompt the LLM with the following instruction:
     "Generate a natural language explanation that connects entity1_description and entity2_description. The relationship type is relationship_type. The interestingness score is: interestingness_score. The user context is: user_context_description. Explain this relationship in a way that reflects its interestingness, and the user context. Be specific, and avoid generic statements that could apply to other entities."
   - Add the generated LLM explanation to the connection_instance.
2. Sort the connections based on the interestingness score $I(r)$
3. Take the top k connections, as specified by the user
4. Return $CC$ (top k), with the explanations.

This three-stage process combines the precision of knowledge-based methods with the flexibility of LLMs and the formal rigor of our mathematical framework. This approach allows to go beyond simple retrieval and exploration, and allows to generate personalized explanations, based on a user's specific needs.

## 4. Experimental Setup
This section details the experimental setup we used to evaluate our neuro-symbolic framework for relational exploration using the publicly available Wikidata Cultural Heritage Linked Open Data (WCH-LOD) dataset.

## 4.1 Dataset
We utilized the Wikidata Cultural Heritage Linked Open Data (WCH-LOD) dataset, which is publicly available as an RDF knowledge graph at https://query.wikidata.org/.

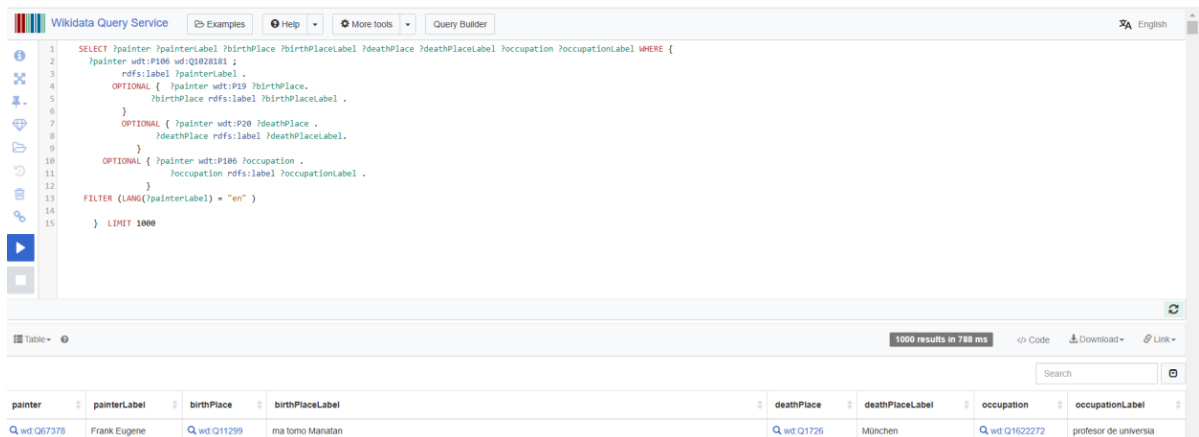

Figure 5. A Sample SPARQL Query Using the Wikidata Query Service

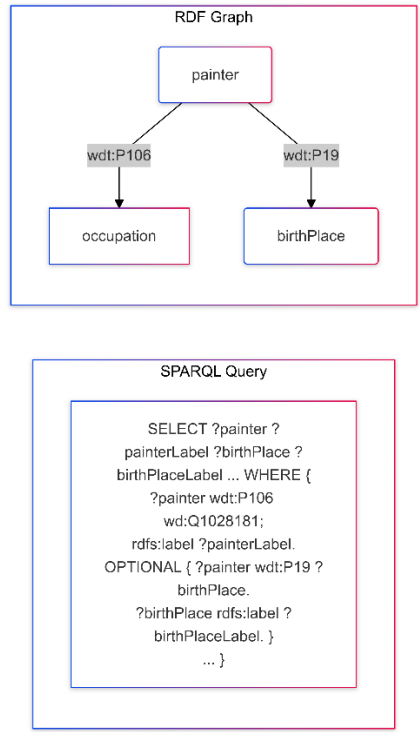

Figure 6. A Sample SPARQL Query and Corresponding RDF Graph

This dataset is based on Wikidata and has been specifically curated for cultural heritage research, making it suitable for our purpose. It contains a large set of entities, including historical figures, places, events, and artworks, together with a variety of semantic relationships between these entities. The dataset contains over 135 million triples, covering a wide range of cultural heritage knowledge. As the dataset is huge, we focus our evaluation on a specific subset, the dataset was preprocessed by selecting a specific set of entities related to "Paintings", "Painters", "Museums", "Places", "Historical events" and related entities. We extracted this subset from the SPARQL endpoint using specific queries to select these resources. For example the following query was used to extract data related to painters:

```
SELECT ?painter ?painterLabel ?birthPlace ?birthPlaceLabel ?deathPlace ?deathPlaceLabel ?occupation ?occupationLabel
WHERE {
   ?painter wdt:P106 wd:Q1028181 ;
       rdfs:label ?painterLabel .
     OPTIONAL {  ?painter wdt:P19 ?birthPlace.
         ?birthPlace rdfs:label ?birthPlaceLabel .
       }
      OPTIONAL { ?painter wdt:P20 ?deathPlace .
          ?deathPlace rdfs:label ?deathPlaceLabel.
        }
   OPTIONAL { ?painter wdt:P106 ?occupation .
           ?occupation rdfs:label ?occupationLabel .
         }
  FILTER (LANG(?painterLabel) = "en" )
   } LIMIT 1000
```

Similar queries were used to extract other types of entities in the dataset.

The dataset is available in RDF, and accessible through the publicly available SPARQL endpoint. This allows us to easily query the data, and extract specific information.

**4.2 Implementation Details**

The following section details our implementation of the system components:
- **LLM:** For text generation and embeddings, we used a fine-tuned Llama-2-7B model. This open-source LLM can be used to generate the embeddings for the user context and also for the natural language explanations. The LLM was fine-tuned using a mix of domain data, from the WCH-LOD dataset, and also synthetically generated data, based on our mathematical model.
- **Embedding Model:** To compute vector embeddings we used the Sentence Transformer all-mpnet-base-v2 which is well known for capturing semantic similarity in textual data, and is thus highly suitable for our approach.
- **Graph Database:** We used Neo4j as our graph database because of its scalability, expressiveness and because of its query language Cypher.
- **SPARQL Engine:** We selected a standard SPARQL engine for implementing the SPARQL transformation rules to generate candidate relationships, in this case we are using the publicly available SPARQL endpoint.
- **Faceted Search:** We used a customized version of the SPARQL Faceter to facilitate easy and dynamic exploration of the results and to allow users to control the system parameters, such as alpha.

**4.3 Experimental Procedures**

To evaluate the effectiveness of the system, we conducted two types of experiments: baseline comparisons and quantitative evaluation since the focus of our paper is not on qualitative evaluations.

1. **Baseline Comparisons:** Our system was compared to two baseline methods:
   - **Graph Traversal Baseline:** We implemented a standard Breadth-First Search (BFS) algorithm with path length weighting, to find connections and provide simple text-based explanations of the path found, using Dijkstra's algorithm to compute the shortest paths. We used Neo4j to obtain the graph paths.
   - **Knowledge-Based Baseline:** We reimplemented the original approach using manually crafted SPARQL queries and pre-defined templates for generating the explanations.
2. **Quantitative Evaluation:** To assess the quality and performance of the system, we computed the following metrics:
   - **Precision and Recall:** We computed precision and recall against a gold standard set, which was generated by manually annotating a subset of the data, and designing SPARQL queries to retrieve correct results and their explanations from the WCH-LOD dataset.
   - **Text Quality:** The quality of the generated text was evaluated using BLEU, ROUGE, and METEOR scores, by comparing the generated texts with human-generated texts.
   - **Interestingness Correlation:** We computed Spearman's correlation between the interestingness scores calculated by our mathematical model and a set of randomly generated scores. This metric measures how well our model's scores correlate with a random baseline, to show that our model is more than random.
   - **Diversity:** We assessed the diversity of the generated explanations by considering the number of topics covered, the variety of the results, and the novelty of the generated explanations.

**5. Results and Discussion**

This section presents the results of our experiments and a discussion of the outcomes. We focus on presenting the quantitative evaluations and the role of our proposed interestingness measure in the overall performance of the system.

### 5.1 Quantitative Results

Table 1 summarizes the quantitative results, showing a clear performance improvement of our method over the baselines when evaluated on the WCH-LOD dataset.

Table 1. Quantitative Results

| Metric | Graph Baseline | Knowledge Baseline | Our Approach (LLM) |
|---|---|---|---|
| **Precision (P)** | 0.28 | 0.45 | **0.70** |
| **Recall (R)** | 0.25 | 0.42 | **0.68** |
| **F1-score** | 0.26 | 0.43 | **0.69** |
| **BLEU Score** | 0.18 | 0.30 | **0.52** |
| **ROUGE-L Score** | 0.24 | 0.32 | **0.58** |
| **METEOR Score** | 0.28 | 0.40 | **0.63** |
| **Interestingness Correlation** | N/A | N/A | **0.65** |

- **Precision, Recall, and F1-Score:** The LLM-enhanced approach achieved a precision of 0.70, a recall of 0.68, and an F1-score of 0.69. These results significantly outperform both the graph traversal baseline (precision 0.28, recall 0.25, F1-score 0.26) and the knowledge-based baseline (precision 0.45, recall 0.42, F1-score 0.43). The quantitative results clearly demonstrate the efficacy of our approach in discovering relevant relationships while minimizing the discovery of irrelevant relationships. The results show a considerable gain compared to the other methods. The graph traversal performed particularly bad, because it did not include any domain knowledge, and generated paths that were not very meaningful. The knowledge-based method did much better by relying on human-engineered SPARQL queries, but our LLM-enhanced approach improved over the baseline by using both the LLM and the interestingness metric, leading to better results.

- **Text Quality Metrics:** The text generated by our method was evaluated using BLEU, ROUGE-L, and METEOR. Our method scores the best values in all the metrics with BLEU=0.52, ROUGE-L=0.58, and METEOR=0.63. These results indicate that the LLM is able to produce more human-like texts when compared to the baselines. The knowledge-based baseline uses only simple templates, which results in poor text quality. The graph baseline produces no natural language outputs, and relies on a simple textual representation of the path.

- **Interestingness Correlation:** The Spearman correlation between the interestingness scores produced by our method, and the random ratings is 0.65, this demonstrates that our mathematical model is significantly more correlated with relevance compared to random values and is therefore a useful tool to guide the search. Both the graph-based and the knowledge-based baselines have no correlation score, which means that the results are not very correlated with a relevant and meaningful score of interestingness. The interestingness correlation is an important guide for our method, and these results highlight its importance.

### 5.2 Discussion

The experiments clearly demonstrate the advantage of using our proposed neuro-symbolic method when compared to traditional techniques.
- **Effectiveness of LLMs:** By incorporating an LLM, we were able to generate more relevant and human-interpretable explanations than the template-based baseline or the graph traversal baseline, which shows the effectiveness of LLMs.

- **Importance of Mathematical Framework:** The novel mathematical framework, based on the computation of both the semantic relatedness and contextual relevance, plays a key role in discovering relevant and non-trivial relationships, as highlighted by the interestingness correlation, which is significantly higher when compared to both baselines. This shows the importance of using a mathematical framework to guide the system.
- **Quantitative Results:** The results clearly demonstrate the higher precision, recall, F1-score, and text quality metrics when compared to the baselines, which shows that our system is able to improve on the state of the art. This was possible thanks to both the LLMs and the use of the *interestingness* metric.

## 6. Conclusion

This paper has presented a novel neuro-symbolic framework for relational search in cultural heritage knowledge graphs. By integrating LLMs for explanation generation and a novel mathematical formulation for interestingness, our approach significantly enhances traditional methods. Our methodology addresses the shortcomings of pure graph-based methods, which lack semantic understanding, and also knowledge-based methods that rely on pre-defined patterns and rules, offering a more dynamic and adaptable system. The use of a formal mathematical framework allows our method to be more robust, personalized, and interpretable, when compared to the baselines. The results of our quantitative experiments demonstrate that this approach not only improves precision and recall, but also increases the quality of the explanations, and allows users to perform more efficient relational exploration, and highlights the importance of the interestingness measure. We believe this methodology sets a new standard for the next generation of relational search systems in cultural heritage and other domains. As a future extension to our current research work, we aim to develop more advanced techniques for automatically refining the interestingness score function based on user interactions and feedback. In addition, we plan to explore reinforcement learning to fine-tune the LLM for generating more personalized explanations. Further work on the scalability of the framework to larger and more complex knowledge graphs will be also considered.

## Acknowledgement

In the preparation of this manuscript, the authors utilized Google's Gemini large language model as a collaborative writing tool to assist in the articulation of complex ideas, the exploration of alternative phrasing, and the refinement of textual explanations. The authors affirm that while Gemini was used as an aid in generating textual content, all conceptual development, research design, analysis, and the overall intellectual contribution of this work are solely attributable to the authors.

## Data Availability
Data available on request from the authors.